\documentclass{article}

\usepackage[final]{nips_2016}

\usepackage[utf8]{inputenc} 
\usepackage[T1]{fontenc}    
\usepackage{hyperref}       
\usepackage{url}            
\usepackage{booktabs}       
\usepackage{amsfonts}       
\usepackage{nicefrac}       
\usepackage{microtype}      
\usepackage{amsfonts}                   
\usepackage{amsmath}                    
\usepackage{graphicx}                   

\title{Lifelong Learning of Spatiotemporal Representations with Dual-Memory Recurrent Self-Organization}

\author{
  German I. Parisi$^1$, Jun Tani$^2$, Cornelius Weber$^1$, Stefan Wermter$^1$\\
  $^1$\small{Knowledge Technology, Department of Informatics, Universit\"at Hamburg, Germany}\\
  $^2$\small{Cognitive Neurorobotics Research Unit, Okinawa Institute of Science and Technology, Japan}\\
}

\begin{document}

\maketitle

\begin{abstract}
\small{Artificial autonomous agents and robots interacting in complex environments are required
to continually acquire and fine-tune knowledge over sustained periods of time. The ability
to learn from continuous streams of information is referred to as lifelong learning and
represents a long-standing challenge for neural network models due to catastrophic
forgetting in which novel sensory experience interferes with existing representations
and leads to abrupt decreases in the performance on previously acquired knowledge.
Computational models of lifelong learning typically alleviate catastrophic forgetting in
experimental scenarios with given datasets of static images and limited complexity,
thereby differing significantly from the conditions artificial agents are exposed to. In
more natural settings, sequential information may become progressively available over
time and access to previous experience may be restricted. Therefore, specialized
neural network mechanisms are required that adapt to novel sequential experience
while preventing disruptive interference with existing representations. In this paper, we
propose a dual-memory self-organizing architecture for lifelong learning scenarios. The
architecture comprises two growing recurrent networks with the complementary tasks
of learning object instances (episodic memory) and categories (semantic memory). Both
growing networks can expand in response to novel sensory experience: the episodic
memory learns fine-grained spatiotemporal representations of object instances in an
unsupervised fashion while the semantic memory uses task-relevant signals to regulate
structural plasticity levels and develop more compact representations from episodic
experience. For the consolidation of knowledge in the absence of external sensory
input, the episodic memory periodically replays trajectories of neural reactivations. We
evaluate the proposed model on the CORe50 benchmark dataset for continuous object
recognition, showing that we significantly outperform current methods of lifelong learning
in three different incremental learning scenarios.}
\end{abstract}

\section{Introduction}

Artificial autonomous agents and robots interacting in dynamic
environments are required to continually acquire and fine-tune
their knowledge over time (Thrun and Mitchell, 1995; Parisi
et al., 2018a). The ability to progressively learn over a sustained
time span by accommodating novel knowledge while retaining
previously learned experiences is referred to as continual or
lifelong learning. In contrast to state-of-the-art deep learning
models that typically rely on the full training set being available
at once (see LeCun et al., 2015 for a review), lifelong learning
systems must account for situations in which the training data
become incrementally available over time. Effective models of
lifelong learning are crucial in real-world conditions where an
autonomous agent cannot be provided with all the necessary
prior knowledge to interact with the environment and the direct
access to previous experience is restricted (Thrun and Mitchell,
1995). Importantly, there may be no distinction between training
and test phases, which requires the system to concurrently
learn and timely trigger behavioral responses (Cangelosi and
Schlesinger, 2015; Tani, 2016).
Lifelong machine learning represents a long-standing
challenge due to catastrophic forgetting or interference,
i.e., training a model with a new task leads to an abrupt
decrease in the performance on previously learned tasks
(McCloskey and Cohen, 1989). To overcome catastrophic
forgetting, computational models must adapt their existing
representations on the basis of novel sensory experience while
preventing disruptive interference with previously learned
representations. The extent to which a system must be flexible
for learning novel knowledge and stable for preventing the
disruption of consolidated knowledge is known as the stability-plasticity dilemma, which has been extensively studied for both
computational and biological systems (e.g., Grossberg, 1980,
2007; Mermillod et al., 2013; Ditzler et al., 2015).
Neurophysiological evidence suggests distributed
mechanisms of structural plasticity that promote lifelong
memory formation, consolidation, and retrieval in multiple
brain areas (Power and Schlaggar, 2016; Zenke et al., 2017a).
Such mechanisms support the development of the human
cognitive system on the basis of sensorimotor experiences
over sustained time spans (Lewkowicz, 2014). Crucially, the
brain must constantly perform two complementary tasks: (i)
recollecting separate episodic events (specifics), and (ii) learning
the statistical structure from the episodic events (generalities).
The complementary learning systems (CLS) theory (McClelland
et al., 1995; Kumaran et al., 2016) holds that these two
interdependent operations are mediated by the interplay of the
mammalian hippocampus and neocortex, providing the means
for episodic memory (specific experience) and semantic memory
(general structured knowledge). Accordingly, the hippocampal
system exhibits quick learning of sparse representations from
episodic experience which will, in turn, be transferred and
integrated into the neocortical system characterized by a slower
learning rate with more compact representations of statistical
regularities.

Re-training a (deep) neural architecture from scratch
in response to novel sensory input can require extensive
computational effort. Furthermore, storing all the previously
encountered data in lifelong learning scenarios has the general
drawback of large memory requirements. Instead, Robins (1995)
proposed pseudo-rehearsal (or intrinsic replay) in which previous
memories are revisited without the need of explicitly storing
data samples. Pseudo-samples are drawn from a probabilistic
or generative model and replayed to the system for memory
consolidation. From a biological perspective, the direct access to
past experiences is limited or restricted. Therefore, the replay of
hippocampal representations in the absence of external sensory
input plays a crucial role in memory encoding (Carr et al.,
2011; Kumaran et al., 2016). Memory replay is argued to
occur through the reactivation of neural patterns during both
sleep and awake states (e.g., free recall; Gelbard-Sagiv et al.,
2008). Hippocampal replay provides the means for the gradual
integration of knowledge into neocortical structures through
the reactivation of recently acquired knowledge interleaved
with the exposure to ongoing episodic experience (McClelland
et al., 1995). Consequently, the periodic replay of previously
encountered samples can alleviate catastrophic forgetting during
incremental learning tasks, especially when the number of
training samples for the different classes is unbalanced or when a
sample is encountered only once (Robins, 1995).

A number of computational approaches have drawn
inspiration from the learning principles observed in biological
systems. Machine learning models addressing lifelong learning
can be divided into approaches that regulate intrinsic levels of
plasticity to protect consolidated knowledge, that dynamically
allocate neural resources in response to novel experience, or
that use complementary dual-memory systems with memory
replay (see section 2). However, most of these methods are
designed to address supervised learning on image datasets of
very limited complexity such as MNIST (LeCun et al., 1998) and
CIFAR-10 (Krizhevsky, 2009) while not scaling up to incremental
learning tasks with larger-scale datasets of natural images and
videos (Kemker et al., 2018; Parisi et al., 2018a). Crucially,
such models do not take into account the temporal structure
of the input which plays an important role in more realistic
learning conditions, e.g., an autonomous agent learning from
the interaction with the environment. Therefore, in contrast to
approaches in which static images are learned and recognized in
isolation, we focus on lifelong learning tasks where sequential
data with meaningful temporal relations become progressively
available over time.

In this paper, we propose a growing dual-memory (GDM)
architecture for the lifelong learning of spatiotemporal
representations from videos, performing continuous object
recognition at an instance level (episodic knowledge) and at a
category level (semantic knowledge). The architecture comprises
two recurrent self-organizing memories that dynamically adapt
the number of neurons and synapses: the episodic memory learns
representations of sensory experience in an unsupervised fashion
through input-driven plasticity, whereas the semantic memory
develops more compact representations of statistical regularities
embedded in episodic experience. For this purpose, the semantic
memory receives neural activation trajectories from the episodic
memory and uses task-relevant signals (annotated labels) to
modulate levels of neurogenesis and neural update. Internally
generated neural activity patterns in the episodic memory
are periodically replayed to both memories in the absence of
sensory input, thereby mitigating catastrophic forgetting during
incremental learning. We conduct a series of experiments
with the recently published Continuous Object Recognition
(CORe50) benchmark dataset (Lomonaco and Maltoni, 2017).
The dataset comprises 50 objects within 10 categories with image
sequences captured under different conditions and containing
multiple views of the same objects (indoors and outdoors,
varying background, object pose, and degree of occlusion). We
show that our model scales up to learning novel object instances
and categories and that it outperforms current lifelong learning
approaches in three different incremental learning scenarios.

\section{Related Work}

The CLS theory (McClelland et al., 1995) provides the basis
for computational frameworks that aim to generalize across
experiences while retaining specific memories in a lifelong
fashion. Early computational attempts include French (1997)
who developed a dual-memory framework using pseudorehearsal (Robins, 1995) to transfer memories, i.e., the training
samples are not explicitly kept in memory but drawn from a
probabilistic model. However, there is no empirical evidence
showing that this or similar contemporaneous approaches (see
O’Reilly and Norman, 2002 for a review) scale up to large-scale
image and video benchmark datasets. More recently, Gepperth
and Karaoguz (2015) proposed two approaches for incremental
learning using a modified self-organizing map (SOM) and a SOM
extended with a short-term memory (STM). We refer to these
two approaches as GeppNet and GeppNet+STM, respectively.
In GeppNet, task-relevant feedback from a regression layer is
used to select whether learning in the self-organizing hidden
layer takes place. In GeppNet+STM, the STM is used to store
novel knowledge which is occasionally played back to the
GeppNet layer during sleep phases interleaved with training
phases. This latter approach yields better performance and faster
convergence in incremental learning tasks with the MNIST
dataset. However, the STM has a limited capacity, thus learning
new knowledge can overwrite old knowledge. In both cases,
the learning process is divided into the initialization and the
actual incremental learning phase. Furthermore, GeppNet and
GeppNet+STM require storing the entire training dataset during
training. Kemker and Kanan (2018) proposed the FearNet model
for incremental class learning inspired by studies of memory
recall and consolidation in the mammalian brain during fear
conditioning (Kitamura et al., 2017). FearNet uses a hippocampal
network capable of immediately recalling new examples, a PFC
network for long-term memories, and a third neural network
inspired by the basolateral amygdala for determining whether
the system should use the PFC or hippocampal network for a
particular example. FearNet consolidates information from its
hippocampal network to its PFC network during sleep phases.
Kamra et al. (2018) presented a similar dual-memory framework
for lifelong learning that uses a variational autoencoder as
a generative model for pseudo-rehearsal. Their framework
generates a short-term memory module for each new task.
However, prior to consolidation, predictions are made using an
oracle, i.e., they know which module contains the associated
memory.

Different methods have been proposed that are based on
regularization techniques to impose constraints on the update
of the neural weights. This is inspired by neuroscience findings
suggesting that consolidated knowledge can be protected from
interference via changing levels of synaptic plasticity (Benna
and Fusi, 2016) and is typically modeled in terms of adding
regularization terms that penalize changes in the mapping
function of a neural network. For instance, Li and Hoiem (2016)
proposed a convolutional neural network (CNN) architecture
in which the network that predicts the previously learned tasks
is enforced to be similar to the network that also predicts the
current task by using knowledge distillation, i.e., the transferring
of knowledge from a large, highly regularized model to a smaller
model. This approach, known as learning without forgetting
(LwF), has the drawbacks of highly depending on the relevance of
the tasks and that the training time for one task linearly increases
with the number of old tasks. Kirkpatrick et al. (2017) proposed
elastic weight consolidation (EWC) which adds a penalty term
to the loss function and constrains the weight parameters that
are relevant to retain previously learned tasks. However, this
approach requires a diagonal weighting over the parameters
of the learned tasks which is proportional to the diagonal of
the Fisher information metric, with synaptic importance being
computed offline and limiting its computational application to
low-dimensional output spaces. Zenke et al. (2017b) proposed to
alleviate catastrophic forgetting by allowing individual synapses
to estimate their importance for solving a learned task. Similar
to Kirkpatrick et al. (2017), this approach penalizes changes to
the most relevant synapses so that new tasks can be learned with
minimal interference. In this case, the synaptic importance is
computed in an online fashion over the learning trajectory in the
parameter space.

In general, regularization approaches comprise additional
loss terms for protecting consolidated knowledge which, with a
limited amount of neural resources, leads to a trade-off on the
performance of old and novel tasks. Other approaches expand
the neural architecture to accommodate novel knowledge. Rusu
et al. (2016) proposed to block any changes to the network trained
on previous knowledge and expand the architecture by allocating
novel sub-networks with a fixed capacity to be trained with the
new information. This prevents catastrophic forgetting but leads
the complexity of the architecture to grow with the number
of learned tasks. Draelos et al. (2017) trained an autoencoder
incrementally using the reconstruction error to show whether
the older digits were retained. Their model added new neural
units to the autoencoder to facilitate the addition of new MNIST
digits. Rebuffi et al. (2017) proposed the iCaRL approach which
stores example data points that are used along with new data
to dynamically adapt the weights of a feature extractor. By
combining new and old data, they prevent catastrophic forgetting
but at the expense of a higher memory footprint.

The approaches described above are designed for the
classification of static images, often exposing the learning
algorithm to training samples in a random order. Conversely,
in more natural settings, we make use of the spatiotemporal
structure of the input. In previous research (Parisi et al., 2017),
we showed that the lifelong learning of action sequences can
be achieved in terms of prediction-driven neural dynamics with
internal representations emerging in a hierarchy of recurrent
self-organizing networks. The networks can dynamically allocate
neural resources and update connectivity patterns according to
competitive Hebbian learning by computing the input based
on its similarity with existing knowledge and minimizing
interference by creating new neurons whenever they are required.
This approach has shown competitive results with batch learning
methods on action benchmark datasets. However, the neural
growth and update are driven by the minimization of the bottomup reconstruction error and, thus, without taking into account
top-down, task-relevant signals that can regulate the plasticitystability balance. Furthermore, the model cannot learn in the
absence of external sensory input, which leads to a non-negligible
degree of disruptive interference during incremental learning
tasks.

\section{Proposed Method}

The proposed architecture with growing dual-memory learning
(GDM) comprises a deep convolutional feature extractor
and two hierarchically arranged recurrent self-organizing
networks (Figure~1). Both recurrent networks are extended
versions of the Gamma-GWR model (Parisi et al., 2017) that
dynamically create new neurons and connections in response
to novel sequential input. The growing episodic memory
(G-EM) learns from sensory experience in an unsupervised
fashion, i.e., levels of structural plasticity are regulated by the
ability of the network to predict the spatiotemporal patterns
given as input. Instead, the growing semantic memory (G-SM)
receives neural activation trajectories from G-EM and uses
task-relevant signals (input annotations) to modulate levels
of neurogenesis and neural update, thereby developing more
compact representations of statistical regularities embedded
in episodic experience. Therefore, G-EM and G-SM mitigate
catastrophic forgetting through self-organizing learning
dynamics with structural plasticity, increasing information
storage capacity in response to novel input.

The architecture classifies image sequences at an instance
level (episodic experience) and a category level (semantic
knowledge). Thus, each input sample carries two labels which
are used for the classification task at the different levels of
the network hierarchy. For the consolidation of knowledge
over time in the absence of sensory input, internally generated
neural activity patterns in G-EM are periodically replayed to
both memories, thereby mitigating catastrophic forgetting during
incremental learning tasks. For this purpose, G-EM is equipped
with synapses that learn statistically significant neural activity
in the temporal domain. As a result, sequence-selective neural
activation trajectories can be generated and replayed after each
learning episode without explicitly storing sequential input.

\begin{figure*}[t]
\centering
\includegraphics[width=0.71\textwidth]{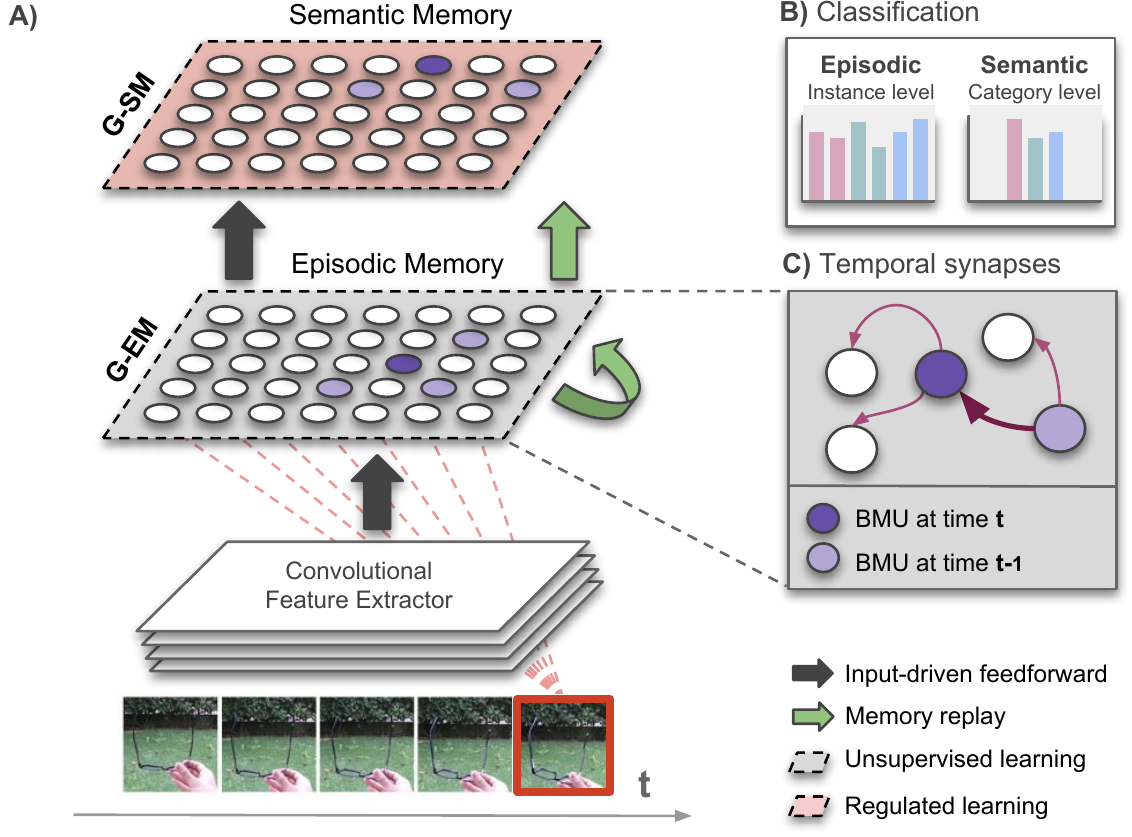}
\caption{(A) Illustration of our growing dual-memory (GDM) architecture for lifelong learning. Extracted features from image sequences are fed into a growing
episodic memory (G-EM) consisting of an extended version of the recurrent Grow-When-Required network (section 3.2). Neural activation trajectories from G-EM are
feed-forwarded to the growing semantic memory (G-SM) that develops more compact representations of episodic experience (section 3.3). While the learning process
of G-EM remains unsupervised, G-SM uses class labels as task-relevant signals to regulate levels of structural plasticity. After each learning episode, internally
generated neural activation trajectories are replayed to both memories (green arrows; section 3.4); (B) The architecture classifies image sequences at instance level
(episodic experience) and at category level (semantic knowledge). For the purpose of classification, neurons in G-EM and G-SM associatively learn histograms of class
labels from the input (red dashed lines); (C) To enable memory replay in the absence of sensory input, G-EM is equipped with temporal synapses that are
strengthened (thicker arrow) between consecutively activated best-matching units (BMU).}
\label{fig:architecture}
\end{figure*}

\subsection{Gamma-GWR}

The Gamma-GWR model~(Parisi et al., 2017) is a recurrent extension of the Grow-When-Required (GWR) self-organizing network (Marsland et al., 2002) that embeds a Gamma memory (Principe et al., 1994) for representing short-term temporal relations.
The Gamma-GWR can dynamically grow or shrink in response to the sensory input distribution.
New neurons will be created to better represent the input and connections (synapses) between neurons will develop according to competitive Hebbian learning, i.e. neurons that activate simultaneously will be connected to each other.
The Gamma-GWR learns the spatiotemporal structure of the input through the integration of temporal context into the computation of the self-organizing network dynamics.

The network is composed of a dynamic set of neurons, $A$, with each neuron consisting of a weight vector $\textbf{w}_j$ and a number $K$ of context descriptors $\textbf{c}_{j,k}$~($\textbf{w}_j,\textbf{c}_{j,k}\in\mathbb{R}^n$).
Given the input $\textbf{x}(t)\in\mathbb{R}^n$, the index of the best-matching unit (BMU), $b$, is computed as:
\begin{equation} \label{eq:GetB}
b = \arg\min_{j\in A}(d_j),
\end{equation}
\begin{equation} \label{eq:BMU}
d_j = \alpha_0 \Vert \textbf{x}(t) - \textbf{w}_j  \Vert^2 + \sum_{k=1}^{K}\ \alpha_k \Vert \textbf{C}_k(t)-\textbf{c}_{j,k}\Vert^2,
\end{equation}
\begin{equation}\label{eq:MergeStep}
\textbf{C}_{k}(t) = \beta \cdot \textbf{w}_b^{t-1}+(1-\beta) \cdot \textbf{c}_{b,k-1}^{t-1},
\end{equation}
where $\Vert \cdot \Vert^2$ denotes the Euclidean distance, $\alpha_i$ and $\beta$ are constant factors that regulate the influence of the temporal context, $\textbf{w}_b^{t-1}$ is the weight vector of the BMU at $t-1$, and $\textbf{C}_{k}\in\mathbb{R}^n$ is the global context of the network with $\textbf{C}_{k}(t_0)=0$.

The activity of the network, $a(t)$, is defined in relation to the distance between the input and its BMU (Equation~\ref{eq:BMU}) as follows:
\begin{equation} \label{eq:Activity}
a(t)=\exp(-d_b),
\end{equation}
thus yielding the highest activation value of $1$ when the network can perfectly predict the input sequence~(i.e.~$d_b=0$).
Furthermore, each neuron is equipped with a habituation counter $h_j \in [0,1]$ expressing how frequently it has fired based on a simplified model of how the efficacy of a habituating synapse reduces over time (Stanley, 1976).
Newly created neurons start with $h_j=1$.
Then, the habituation counter of the BMU, $b$, and its neighboring neurons, $n$, iteratively decrease towards 0.
The habituation rule (Marsland et al., 2002) for a neuron $i$ is given by:
\begin{equation}\label{eq:FiringCounter}
\Delta h_i=\tau_i \cdot \kappa \cdot (1-h_i)-\tau_i,
\end{equation}
with $i\in\{b,n\}$ and where $\tau_i$ and $\kappa$ are constants that control the monotonically decreasing behavior.
Typically, $h_b$ is decreased faster than $h_n$ with $\tau_b>\tau_n$.

The network is initialized with two neurons and, at each learning iteration, a new neuron is created whenever the activity of the network, $a(t)$, in response to the input $\textbf{x}(t)$ is smaller than a given insertion threshold $a_T$.
Furthermore, $h_b$ must be smaller than a habituation threshold $h_T$ in order for the insertion condition to hold, thereby fostering the training of existing neurons before new ones are added.
The new neuron is created halfway between the BMU and the input.
The training of the neurons is carried out by adapting the BMU $b$ and the neurons $n$ to which the $b$ is connected:
\begin{equation}\label{eq:UpdateRateW}
\Delta \textbf{w}_i = \epsilon_i \cdot h_i \cdot (\textbf{x}(t) - \textbf{w}_i),
\end{equation}
\begin{equation}\label{eq:UpdateRateC}
\Delta \textbf{c}_{i, k} = \epsilon_i \cdot h_i \cdot (\textbf{C}_k(t) - \textbf{c}_{i, k}),
\end{equation}
with $i\in\{b,n\}$ and where $\epsilon_i$ is a constant learning rate ($\epsilon_n<\epsilon_b$).
Furthermore, the habituation counters of the BMU and the neighboring neurons are updated according to Equation~\ref{eq:FiringCounter}.
Connections between neurons are updated on the basis of neural co-activation, i.e. when two neurons fire together (BMU and second-BMU), a connection between them is created if it does not yet exist.
Each connection has an age that increases at each learning iteration.
The age of the connection between the BMU and the second-BMU is reset to $0$, whereas the other ages are increased by a value of $1$.
The connections with an age greater than a given threshold can be removed, and neurons without connections can be deleted.

For the purpose of classification, an associative matrix $H(j,l)$ stores the frequency-based distribution of sample labels during the learning phase so that each neuron $j$ stores the number of times that an input with label $l$ had $j$ as its BMU.
Thus, the predicted label $\xi_j$ for a neuron $j$ can be computed as:
\begin{equation}\label{eq:WinnerLabel}
\xi_j = \arg\max_{l \in L} H(j,l),
\end{equation}
where $l$ is an arbitrary label.
Therefore, the unsupervised Gamma-GWR can be used for classification without requiring the number of label classes to be predefined.

\subsection{Episodic Memory}

The learning process of growing episodic memory G-EM is unsupervised, thereby creating new neurons or updating existing ones to minimize the discrepancy between the sequential input and its neural representation.
In this way, episodic memories can be acquired and fine-tuned iteratively through sensory experience.
This is functionally consistent with hippocampal representations, e.g. in the dentate gyrus, which are responsible for pattern separation through the orthogonalization of incoming inputs supporting the auto-associative storage and retrieval of item-specific information from individual episodes~(Yassa and Stark, 2011; Neuneubel and Knierim, 2014). 

Given an input image frame, the extracted image feature vector (see section~4.1) is given as input to G-EM which recursively integrates the temporal context into the self-organizing neural dynamics.
The spatial resolution of G-EM neurons can be tuned through the insertion threshold, $a_T$, with a greater $a_T$ leading to more fine-grained representations since new neurons will be created whenever $a(t)<a_T$ (see Equation~\ref{eq:Activity}).
The temporal depth is set by the number of context descriptors, $K$, with a greater $K$ yielding neurons that activate for larger temporal windows (longer sequences), whereas the temporal resolution is set by the hyperparameters $\alpha$ and $\beta$~(see Equation~\ref{eq:BMU} and \ref{eq:MergeStep}).

To enable memory replay in the absence of external sensory input, we extend the Gamma-GWR model by implementing temporal connections that learn trajectories of neural activity in the temporal domain.
Such temporal connections are sequence-selective synaptic links which are incremented between two consecutively activated neurons~(Parisi et al., 2016).
Sequence selectivity driven by asymmetric connections has been argued to be a feature of the cortex (Mineiro and Zipser, 1998), where an active neuron pre-activates neurons encoding future patterns.
Formally, when the two neurons $i$ and $j$ are consecutively activated at time $t-1$ and $t$ respectively, their temporal synaptic link $P_{(i,j)}$ is increased by $\Delta P_{(i,j)} = 1$.
For each neuron $i\in A$, we can retrieve the next neuron $v$ of a prototype trajectory by selecting
\begin{equation}\label{eq:Next}
v=\arg\max_{j \in A \setminus {i}} P_{(i,j)}.
\end{equation}
Recursively generated neural activation trajectories can be used for memory replay~(see section 3.4).
During the learning phase, G-EM neurons will store instance-level label classes $\xi^I$ for the classification of the input~(see~Equation~\ref{eq:WinnerLabel}).
Furthermore, since trajectories of G-EM neurons are replayed to G-SM in the absence of sensory input, G-EM neurons will also store labels at a category label $l^C$.
Therefore, the associative matrix for each neuron $j$ is of the form $H(j,l^I,l^C)$.

\subsection{Semantic Memory}

The growing semantic memory G-SM combines bottom-up drive from neural activity in G-EM and top-down signals (i.e. category-level labels from the input) to regulate structural plasticity levels.
More specifically, the mechanisms of neurogenesis and neural weight update are regulated by the ability of G-SM to correctly classify its input.
Therefore, while G-EM iteratively minimizes the discrepancy between the input sequences and their internal representations, G-SM will create new neurons only if the correct label of a training sample cannot be predicted by its BMU in G-SM.
This is implemented as an additional constraint in the condition for neurogenesis so that new neurons are not created unless the predicted label of the BMU~(Equation~\ref{eq:WinnerLabel}) does not match the input label.

G-SM receives as input activated neural weights from G-EM, i.e. the weight vector of a BMU in G-EM, $\textbf{w}_b^{\text{EM}}$, for a given input frame.
As an additional mechanism to prevent novel sensory experience from interfering with consolidated representations, G-SM neurons are updated (Equation~\ref{eq:UpdateRateW} and \ref{eq:UpdateRateC}) only if the predicted label for the BMU in G-SM matches in the input label, i.e. if the BMU codes for the same object category as the input.
In this way, the representations of an object category cannot be updated in the direction of the input belonging to a different category, which would cause disruptive interference.

As a result of hierarchical processing, G-SM neurons code for information acquired over larger temporal windows than neurons in G-EM.
That is, one G-SM will fire for a number $K^{\text{SM}}+1$ of neurons fired in G-EM (where $K^{\text{SM}}$ is the temporal depth of G-SM neurons).
Since G-EM neurons will fire for a number $K^{\text{EM}}+1$ of input frames, G-SM neurons will code for a total of $K^{\text{SM}}+K^{\text{EM}}+1$ input frames.
This is consistent with established models of memory consolidation where neocortical representations code for information acquired over more extended time periods than the hippocampus  (e.g.,
Kumaran and McClelland, 2012; Kumaran et al., 2016), thereby
yielding a higher degree of temporal slowness

Temporal slowness results from the statistical learning of
spatiotemporal regularities, with neurons coding for prototype
sequences of sensory experience. By using category-level signals
to regulate neural growth and update, G-SM will develop more
compact representations from episodic experience with neurons
activating in correspondence of semantically-related input, e.g.,
the same neuron may activate for different instances of the same
category and, because of the processing of temporal context,
the same object seen from different angles. However, specialized
mechanisms of slow feature analysis can be implemented that
would yield invariance to complex input transformations such
as view invariance (e.g., Berkes and Wiskott, 2005; Einhäuser
et al., 2005). View invariance of objects is a prominent property
of higher-level visual areas of the mammalian brain, with neurons
coding for abstract representations of familiar objects rather than
for individual views and visual features (Booth and Rolls, 1998;
Karimi-Rouzbahani et al., 2017). Neurophysiological studies
evidence that distributed representations in high-level visual
regions of the neocortex (semantic) are less sparse than those
of the hippocampus (episodic) and where related categories are
represented by overlapping neural codes (Clarke and Tyler, 2014;
Yamins et al., 2018).

\subsection{Memory Replay}

Hippocampal replay provides the means for the gradual
integration of knowledge into neocortical structures and
is thought to occur through the reactivation of recently
acquired knowledge interleaved with the exposure to ongoing
experiences (McClelland et al., 1995). Although the periodic
replay of previous data samples can alleviate catastrophic
forgetting, storing all previously encountered data samples has
the general drawback of large memory requirements and large
retraining computational times.

In pseudo-rehearsal (or intrinsic replay), memories are drawn
from a probabilistic or generative model and replayed to the
system for memory consolidation (Robins, 1995). In our case,
however, we cannot simply draw or generate isolated and
randomly selected pseudo-samples from a given distribution
since we must account for preserving the temporal structure
of the input. Therefore, we generate pseudo-patterns in terms
of temporally-ordered trajectories of neural activity. For this
purpose, we propose to use the asymmetric temporal links of
G-EM (section 3.2) to recursively reactivate sequence-selective
neural activity trajectories (RNATs) embedded in the network.
RNATs can be computed for each neuron in G-EM for a given
temporal window and replayed to G-EM and G-SM after each
learning episode triggered by external input stimulation.

For each neuron $j$ in G-EM, we generate a RNAT, $S_j$, of length $\lambda=K^{\text{EM}}+K^{\text{SM}}+1$ as follows:
\begin{equation}\label{eq:RNATs}
S_j=\langle \textbf{w}^{\text{EM}}_{\text{s}(0)},\textbf{w}^{\text{EM}}_{\text{s}(1)},...,\textbf{w}^{\text{EM}}_{\text{s}(\lambda)} \rangle,
\end{equation}
\begin{equation}\label{eq:RNATs1}
\text{s}(i) = \arg\max_{n \in A \setminus {j}} P_{(n,\text{s}(i-1))}, i \in [1,\lambda],
\end{equation}
where $P_{(i,j)}$ is the matrix of temporal synapses~(as defined by Equation~\ref{eq:Next}) and $\text{s}(0)=j$.
The class labels of the pseudo-patters in $S_j$ can be retrieved according to Equation~\ref{eq:WinnerLabel}.

The set of generated RNATs from all G-EM neurons is
replayed to G-EM and G-SM after each learning episode, i.e., a
learning epoch over a batch of sensory observations. As a result
of computing RNATs, sequence-selective prototype sequences
can be generated and periodically replayed without the need of
explicitly storing the temporal relations and labels of previously
seen training samples. This is conceptually consistent with
neurophysiological studies evidencing that hippocampal replay
consists of the reactivation of previously stored patterns of neural
activity occurring predominantly after an experience (Kudrimoti
et al., 1999; Karlsson and Frank, 2009).

\section{Experimental Results}

We perform a series of experiments evaluating the performance
of the proposed GDM model in batch learning (section 4.2),
incremental learning (section 4.3), and incremental learning
with memory replay (section 4.4). We analyze and evaluate
our model with the CORe50 dataset (Lomonaco and Maltoni,
2017; see section 4.1), a recently published benchmark for
continuous object recognition from video sequences. We
reproduce three experimental conditions defined by the CORe50
benchmark (section 4.5) showing that our model significantly
outperforms state-of-the-art lifelong learning approaches. For
the replication of these experiments, the source code of the GDM
model is available as a repository.\footnote{GDM model: https://github.com/giparisi/GDM}

\subsection{Feature Extraction}

\begin{figure*}[t]
\centering
\includegraphics[width=\textwidth]{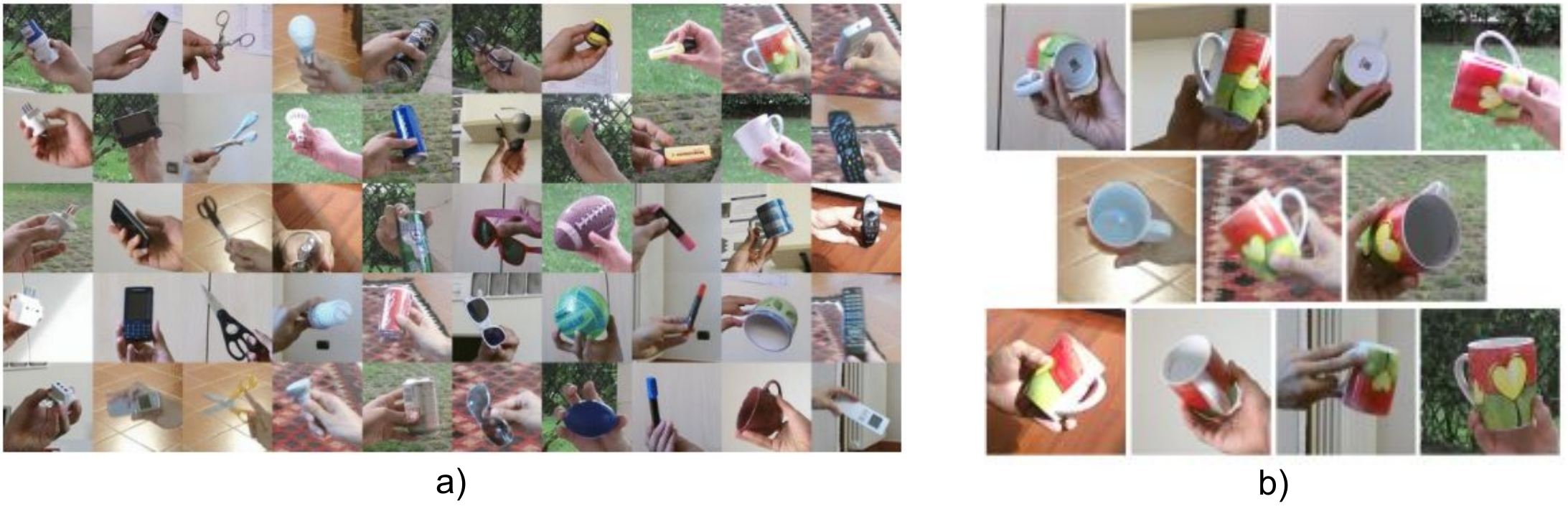}
\caption{The CORe50 dataset designed for continuous object recognition: (A) Example frames of the 10 categories (columns) comprising 5 object instances each,
(B) Example frames for one object instance from the 11 acquisition sessions showing different background, illumination, pose, and degree of occlusion. Adapted from
Lomonaco and Maltoni (2017).}
\label{fig:core50}
\end{figure*}

The CORe50 comprises 50 objects within 10 categories with image sequences captured under different conditions and multiple views of the same objects (varying background, object pose, and degree of occlusion; see Figure~\ref{fig:core50}).
Each object comprises a video sequence of approximately 15 seconds where the object is shown to the vision sensor held by a human operator.
The video sequences were collected in 11 sessions (8 indoors, 3 outdoors) with a Kinect 2.0 sensor delivering RGB ($1027 \times 575$) and depth images ($512 \times 242$) at 20 frames per second~(fps) for a total of 164,866 frames.
For our experiments, we used $128 \times 128$ RGB images provided by the dataset at a reduced frame rate of 5hz.
The movements performed by the human operator with the objects (e.g. rotation) are quite smooth and reducing the number of frames per second has not shown significant loss of information.

For a more direct comparison with the baseline results provided by Lomonaco and Maltoni (2017) who adopted the VGG model (Simonyan and Zisserman, 2014) pre-trained on the ILSVRC-2012 dataset~(Russakovsky et al., 2014), our feature extraction module consists of the same pre-trained VGG model to which we applied a convolutional operation with 256 1x1 kernels on the output of the fully-connected hidden layer to reduce its dimensionality from 2048 to 256.
Therefore, G-EM receives a 256-dimensional feature vector per sequence frame.
Such compression of the feature vectors is desirable since the Gamma-GWR uses the Euclidean distance as a metric to compute the BMUs, which becomes weakly discriminant when the data are very high-dimensional or sparse~(Parisi et al., 2015). 
Furthermore, it is expected that different pre-trained models may exhibit a slightly better performance than VGG, e.g. ResNet-50 (He et al., 2016; see Lomonaco and Maltoni, 2018 for ResNet-50 performance on CORe50).
However, here we focus on showing the contribution of context-aware growing networks rather than comparing deep feature extractors.

\subsection{Batch Learning}

We trained the architecture on the entire training data at once and subsequently tested its classification performance at instance and category level.
Following the same evaluation scheme described by Lomonaco and Maltoni (2017), we used the samples from sessions $\#3$, $\#7$, $\#10$ for testing and the samples from the remaining 8 sessions for training.
We compare our results to the baseline provided by Lomonaco and Maltoni (2017) using fine-tuning on a pre-trained VGG network (VGG+FT).
To better assess the contribution of temporal context (TC) for the task of continuous object recognition, we performed batch learning experiments with 3 different model configurations:
\begin{itemize}
  \item \textbf{GDM}: We trained the model using TC and tested it on novel sequences. For each input frame, an object instance and an object category are predicted.
  \item \textbf{GDM (No TC)}: We trained and tested the model without TC by setting $K=0$, i.e. the computation of the BMU is reduced to $b = \arg\min_{j\in A}\Vert \textbf{x}(t) - \textbf{w}_j  \Vert^2$.
    \item \textbf{GDM (No TC during test)}: We trained the model with TC but tested on single image frames by setting $K=0$ during the test phase.
\end{itemize}

The training hyperparameters are listed in Table~\ref{Tab:parameters}.
Except for the insertion thresholds $a^{\text{EM}}_T$ and $a^{\text{SM}}_T$, the remaining parameters were set similar to Parisi et al. (2017) for the incremental learning of sequences.
Larger insertion thresholds will lead to a larger number of neurons.
However, the best classification performance will not be necessarily obtained by the largest number of neurons.
In G-EM, the neural representation should be characterized by a sufficiently high spatiotemporal resolution for discriminating between similar object instances and replaying episodic experience in the absence of sensory input.
Conversely, regulated unsupervised learning in G-SM will lead to a more compact, overlapping neural representation with a smaller number of neurons while preserving the ability to correctly classify its input.
The number of context descriptors ($K^{\text{EM}}$, $K^{\text{SM}}$) is set to $2$.
This means that G-EM neurons will activate in correspondence of 3 image frames and G-SM neurons in correspondence of 3 G-EM neurons, i.e. a processing window of 5 frames (1s of video at 5fps).
Additional experiments showed that increasing the number of context descriptors does not significantly improve the overall accuracy.
This is because a small number of context descriptors will lead to learning short-term temporal relations which are useful for temporal slowness, i.e. neurons that activate for multiple similar views of the same object (where different views of the object are induced by object motion).
Neurons with a higher temporal depth will learn longer-term temporal relations and, depending on the difference between the training and test set, training with longer sequences may result in the specialization of neurons to the sequences in the training set while failing to generalize.
Therefore, convenient values for $K^{\text{EM}}$ and $K^{\text{SM}}$ can be selected according to different criteria and properties of the input, e.g. number of frames per second, smoothness of object motion, desired degree of neural specialization.

The classification performance for the 3 different configurations is summarized in Table \ref{Tab:comparison}, showing instance-level and category-level accuracy after 35 training epochs averaged across 5 learning trials in which we randomly shuffled the batches from different sessions.
The best results were obtained by GDM using temporal context with an average accuracy of 79.43\% (instance level) and 93.92\% (category level), showing an improvement of 10.35\% and 13.69\% respectively with respect to the baseline results (Lomonaco and Maltoni, 2017).
Without the use of temporal context, the accuracy is comparable to the baseline showing a marginal improvement of 1.34\% (instance level) and 3.31\% (category level).
Our results demonstrate that learning the temporal relations of the input plays an important role for this dataset.
Interestingly, dropping the temporal component during the test phase, i.e. using single image frames for testing on context-aware networks, shows a slightly better performance (2.14\% and 3.78\% respectively) than training without temporal context.
This is because trained neural weights embed some temporal structure of the training sequences and, consequently, the context-free computation of a BMU from a single input frame will still be matched to context-aware neurons.

\begin{table}[t!]
\caption{Training hyperparameters for the G-EM and G-SM networks (batch and incremental learning).}
\label{Tab:parameters}
\begin{center}
\begin{tabular}{llcccc}\toprule
\textbf{Hyperparameters} & \textbf{Value} \\\midrule
Insertion thresholds & $a^{\text{EM}}_T=0.3$, $a^{\text{SM}}_T=0.001$ \\
Habituation counters & $h_T=0.1$, $\tau_{b}=0.3$, $\tau_{n}=0.1$, $\kappa=1.05$ \\
Temporal depth & $K^{\text{EM}}= 2$, $K^{\text{SM}}= 2$ \\
Temporal context & $\alpha=[0.67,0.24,0.09]$, $\beta=0.7$ \\
Learning rates & $\epsilon_b=0.5$, $\epsilon_n=0.005$ \\\midrule
\end{tabular}
\end{center}
\end{table}

\begin{table}[t!]
\caption{Comparison of batch learning performance for instance-level and category-level classification. We show the accuracy for the pre-trained VGG with fine-tuning (VGG+FT) and the proposed GDM for three different configurations: (i) growing networks with temporal context (TC), (ii) without TC, and (iii) without TC during test. Best results in bold.}
\label{Tab:comparison}
\begin{center}
\begin{tabular}{lccccc}\toprule
& \textbf{Accuracy (\%)} & \textbf{Accuracy (\%)} \\
\textbf{Approach} & (Instances) & (Categories) \\\midrule
VGG + FT (Lomonaco and Maltoni, 2017) & 69.08 & 80.23 \\
Proposed GDM (No TC) & 70.42 & 83.54 \\
Proposed GDM (No TC during test) & 72.56 & 87.32 \\
Proposed GDM & \textbf{79.43} & \textbf{93.92} \\\midrule
\end{tabular}
\end{center}
\end{table}

\begin{figure*}[t]
\centering
\includegraphics[width=\textwidth]{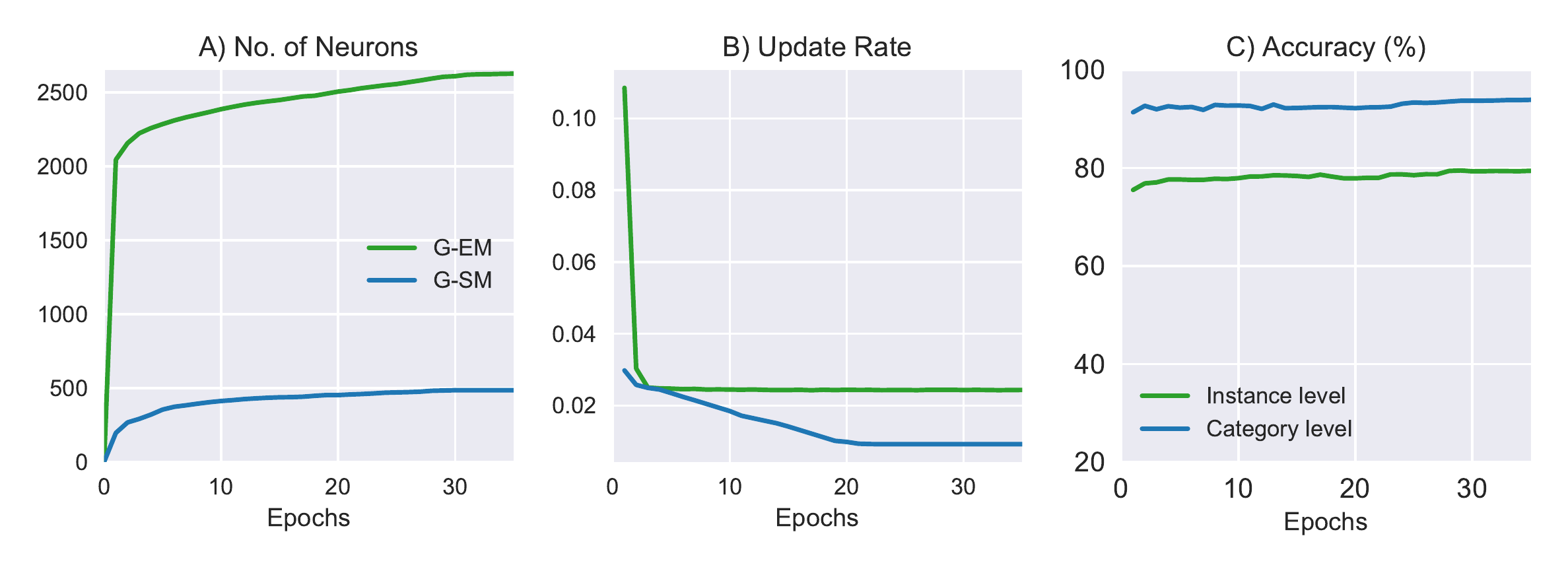}
\caption{Batch learning on the CORe50: numbers of neurons (A), update rates (B), and classification accuracies (C) of G-EM and G-SM through 35 training
epochs averaged across 5 learning trials.}
\label{fig:core50batch}
\end{figure*}

Figure~\ref{fig:core50batch} shows the number of neurons, update rate, and classification accuracy for G-EM and G-SM (with temporal context) through 35 training epochs averaged across 5 learning trials.
It can be seen that the average number of neurons created in G-EM is significantly higher than in G-SM~(Figure~\ref{fig:core50batch}.A).
This is expected since G-EM will grow to minimize the discrepancy between the input and its internal representation, whereas neurogenesis and neural update rate in G-SM are regulated by the ability of the network to predict the correct class labels of the input.
The update rate~(Figure~\ref{fig:core50batch}.B) is given by multiplying the fixed learning rate by the habituation counter of the neurons ($\epsilon_i \cdot h_i$), which shows a monotonically decreasing behavior.
This indicates that, after a number of epochs, the created neurons become habituated to the input.
Such a habituation mechanism has the advantage of protecting consolidated knowledge from being disrupted or overwritten by the learning of novel sensory experience, i.e. well-trained neurons will respond slower to changes in the distribution and the network will create new neurons to compensate for the discrepancy between the input and its representation.

\begin{figure*}[t]
\centering
\includegraphics[width=\textwidth]{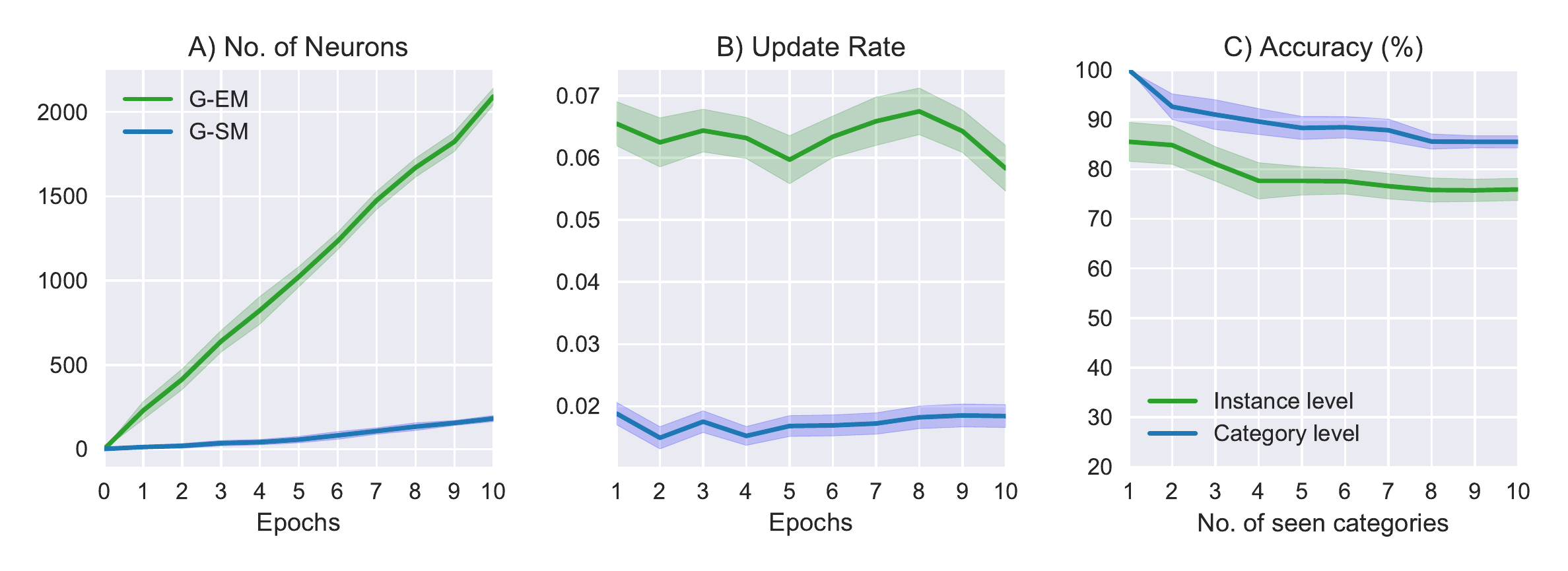}
\caption{Incremental learning: numbers of neurons (A), update rates (B), and classification accuracies (C) over 10 categories averaged across 5 learning trials.
The shaded areas show the standard deviation.}
\label{fig:core50incremental}
\end{figure*}

\subsection{Incremental Learning}

In the incremental learning strategy, the training samples of
different object categories become progressively available over
time, i.e., each mini-batch contains all the instances of an object
category from all the 8 training sessions. Each category batch is
shown once to the model and samples from that category are not
shown again during the learning of new categories. Therefore,
the model must incrementally learn new object instances and
categories without forgetting previously learned ones. For a direct
comparison with our previous experiment, the hyperparameters
for the incremental learning experiment are the same as for the
batch learning strategy~(Table~\ref{Tab:parameters}).

Figure~\ref{fig:core50incremental} shows the number of neurons, update rate, and accuracy over 10 epochs (i.e. the 10 object categories) averaged across 5 runs of randomly shuffled object categories.
The variance from the mean values (shaded areas in Figure~\ref{fig:core50incremental}) shows that the order of exposure to object categories can affect the final result.
In general, the number of neurons increases over time (Figure~\ref{fig:core50incremental}.A) and, in contrast to the batch learning strategy where neurogenesis is particularly strong during the initial training epochs~(Figure~\ref{fig:core50batch}.A), in this case new neurons are progressively created in response to the exposure of the model to novel object classes.
Similarly, the update rate for both networks (Figure~\ref{fig:core50incremental}.B) does not monotonically decrease over time but rather stays quite stable in correspondence to novel sensory experience.
Since newly created neurons are not well trained, the update rate will be higher at the moment of neural insertion and progressively decrease as the newly created neurons become habituated.
The overall accuracy decreases with the number of object categories encountered, showing a higher sensitivity of the model with respect to the order in which the object categories are presented~(Figure~\ref{fig:core50incremental}.C).
The average classification accuracy for the incremental learning strategy is $75.93\% \pm 2.23$ (instance level) and $85.53\% \pm 1.35$ (category level), showing a decrease of $3.5\%$ and $8.39\%$ respectively compared to the batch learning performance.
This suggests that an additional mechanism such as memory replay is required to prevent the disruptive interference of existing representations.

Figure~\ref{fig:core50replay} shows a comparison of the effects of forgetting during the incremental learning strategy in terms of the overall accuracy on the categories encountered so far and the accuracy on the first encountered category as new categories are learned.
For the object instances, we compare the overall accuracy (Figure~\ref{fig:core50replay}.A) with the accuracy on the first 5 encountered instances (i.e. 1 category; see Figure~\ref{fig:core50replay}.B), showing that for the latter the accuracy drops to $69.25\% \pm 4.31$ (compared to $75.93\% \pm 2.23$).
For the object categories~(Figure~\ref{fig:core50replay}.C-D), the accuracy on the first encountered category drops to $79.53\% \pm 5.23$ (compared to $85.53\% \pm 1.35$).
Overall, these results suggest that memory replay is an important feature for the reactivation of previously learned neural representations at the moment of learning from novel sensory experience with the goal to prevent that classes that have been encountered at early stages be forgotten over time.

\subsection{Incremental Learning with Memory Replay}

In this learning strategy, we trained the model as described above with progressively available mini-batches containing 1 object category each.
Here, however, after each learning episode (i.e. a training epoch over the mini-batch), the model generates a set of RNATs, $S_j$ (Equation~\ref{eq:RNATs} and \ref{eq:RNATs1}) from the G-EM neurons.
Thus, the number of RNATs of length $\lambda=5$ is equal to the number of neurons created by G-EM.
The set of RNATs is replayed to G-EM and G-SM in correspondence of novel sensory experience to reinforce previously encountered categories.
Since the growing self-organizing networks store the global temporal context, $\textbf{C}_{k}(t)$, over the training iterations (Equation~\ref{eq:MergeStep}) for learning the temporal structure of the input, each RNAT is fed into G-EM and G-SM as a single sample batch and the global temporal context is reset to zero after one epoch.
It is expected that, by periodically replaying RNATs when new categories are encountered, knowledge representations will consolidate over time and, consequently, significantly alleviate catastrophic forgetting.

\begin{figure*}[t]
\centering
\includegraphics[width=0.85\textwidth]{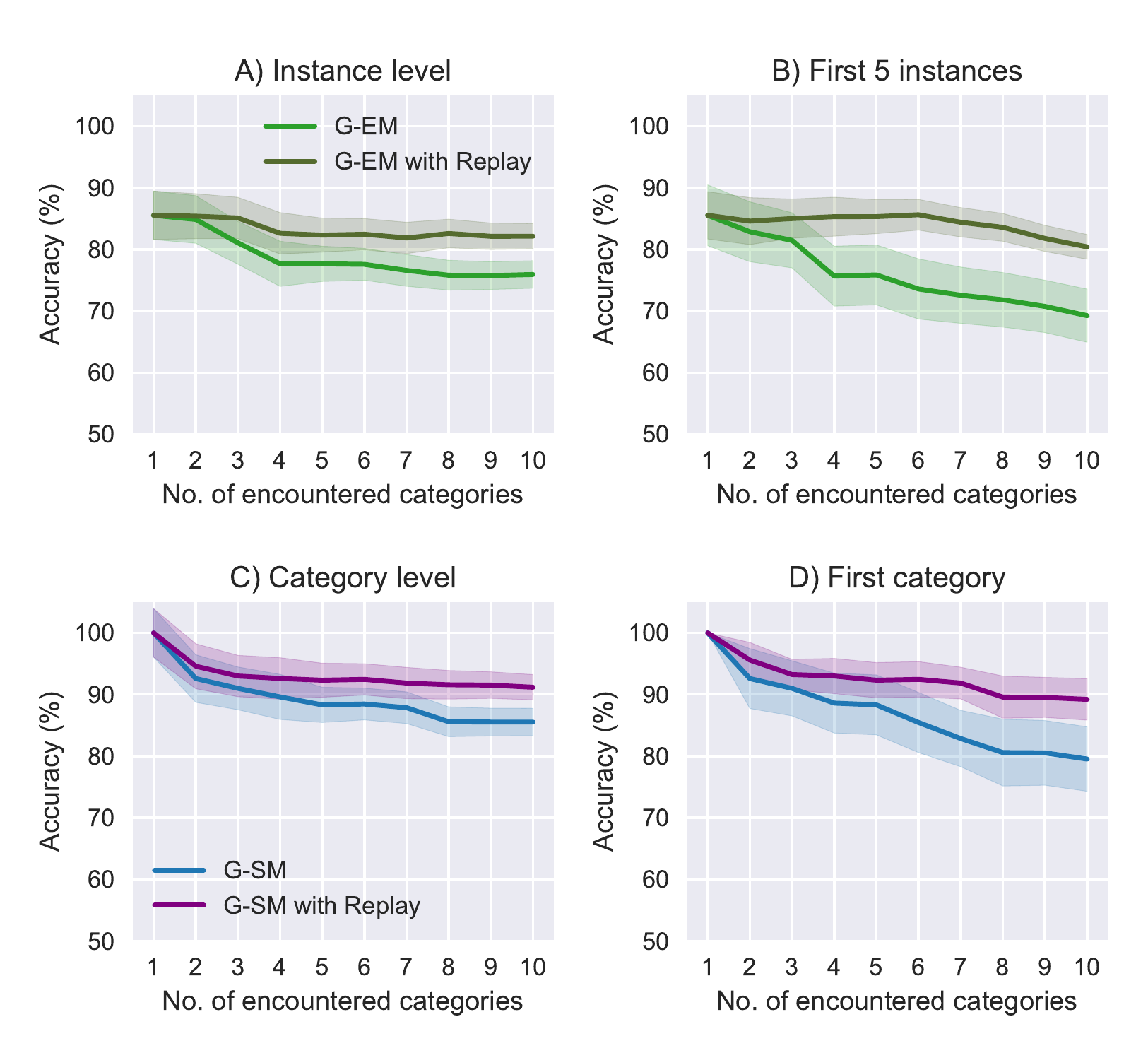}
\caption{Comparison of the effects of forgetting during incremental learning with and without memory replay at an instance level (A,B) and category level (C,D).
Each category contains 5 instances. The plots show the average accuracies on the categories encountered so far (A,C) and the accuracies on the first encountered
category (B,D) as further new categories are learned. The shaded areas show the standard deviation.}
\label{fig:core50replay}
\end{figure*}

The benefit of using memory replay is shown in Figure~\ref{fig:core50replay} where we compare the overall accuracy on all the categories encountered so far to the accuracy on the first encountered category over the number of encountered categories.
At an instance level~(Figure~\ref{fig:core50replay}.A-B), incremental learning with memory replay improves the overall accuracy to $82.14\% \pm 2.05$ (from $75.93\% \pm 2.23$) and accuracy on the first 5 instances to $80.41\% \pm 1.35$ (from $69.25\% \pm 2.01$).
At a category level~(Figure~\ref{fig:core50replay}.C-D), the overall accuracy increases to $91.18\% \pm 0.25$ (from $85.53\% \pm 1.35$) and the accuracy on the first encountered category to $89.21\% \pm 3.37$ (from $79.53\% \pm 5.23$)
Overall, our results support the hypothesis that replaying RNATs generated from G-EM mitigates the effects of catastrophic forgetting.

\subsection{Continuous Object Recognition}

We evaluate our model with the 3 incremental learning scenarios proposed by the CORe50 benchmark for the task of continuous object recognition:

\textbf{New Instances (NI)}: New instances of the same class and from different acquisition sessions become progressively available and are shown once to the model.
Therefore, all the classes to be learned are known.
For all the classes, the model is trained with the instances of a first session and subsequently with the remaining 7 sessions. (Here, the term \textit{classes} is used for object categories.)

\textbf{New Classes (NC)}: Training samples from novel different classes become available over time, thus the model must deal with the learning of new classes without forgetting previously learned ones.
Each training batch contains all the sequences of a small group of classes and memory replay is possible across batches.
The first batch includes 10 objects while the remaining 8 batches contain 5 objects each.
The test set includes samples from all the classes and the model is required to classify samples that have not been seen yet (except for the last evaluation step).

\begin{table}[t!]
\caption{Accuracy on the CORe50 incremental learning scenarios. Results denoted with * indicate the re-implementation of the method by Lomonaco and
Maltoni (2017). Best results in bold.}
\label{Tab:nics}
\begin{center}
\begin{tabular}{llcccc}\toprule

\textbf{Method} & \textbf{Avg. Acc. (\%)} & \textbf{Std. Dev. (\%)} \\\midrule
\textit{New Instances (NI)}  \\\midrule
Proposed GDM (with replay) & \textbf{87.94} & 1.72\\
Proposed GDM & 74.87 & 2.54 \\
Cumulative (Lomonaco and Maltoni, 2017) & 65.15 & 0.66\\
LwF (Zhizhong and Hoiem, 2016) & 59.42 * & 2.71\\
EWC (Kirkpatrick et al., 2017) & 57.40 * & 3.80\\
Na\"ive (Lomonaco and Maltoni, 2017) & 54.69 & 6.18\\\midrule
\textit{New Classes (NC) } &  \\\midrule
Proposed GDM (with replay) & \textbf{86.14} & 2.03\\
Proposed GDM & 73.02 & 2.91 \\
Cumulative & 64.65 & 1.04\\
iCaRL (Rebuffi et al., 2016) & 43.62 * & 0.66\\
CWR (Lomonaco and Maltoni, 2017) & 42.32 & 1.09\\
LwF & 27.60 * & 1.70\\
EWC & 26.22 * & 1.18\\
Na\"ive & 10.75 & 0.84\\\midrule
\textit{New Instances and Classes (NIC) } &  \\\midrule
Proposed GDM (with replay) & \textbf{87.06} & 2.13\\
Proposed GDM & 72.57 & 2.96 \\
Cumulative & 64.13 & 0.88\\
CWR & 29.56 & 0\\
LwF & 28.94 * & 4.30\\
EWC & 28.31 * & 4.30\\
Na\"ive & 19.39 & 2.90\\\midrule
\end{tabular}
\end{center}
\end{table}

\textbf{New Instances and Classes (NIC)}: New instances and classes become available over time, requiring the model to consolidate knowledge about known classes and to learn new ones.
The first batch includes 10 classes and the subsequent batches 5 classes each, with only one training sequence per class included in the batches.
This scenario comprises 79 batches, maximizing the categorical representation in the first batch and randomly selecting the remaining 78 batches.

For each scenario, we compute the average accuracy over 10 configurations of randomly shuffled batches.
The results for the NI, NC, and NIC scenarios compared to other approaches are listed in Table ~\ref{Tab:nics}.
It can be seen that our proposed method with memory replay produces state-of-the-art results for this benchmark dataset, showing an average accuracy of 87,94\%, 86.14\%, and 87.06\% for the NI, NC, and NIC scenarios, respectively.
These results represent a large increase in accuracy over 20\% for each scenario with respect to the previous best results, i.e., a cumulative approach reported by Lomonaco and Maltoni (2017).
The authors reported results using 3 methods with pre-trained CNN models and $128\times128$ images: (i) a na\"ive approach which consists of continuous stochastic gradient descent training as new batches become available, (ii) a proposed \textit{CopyWeights with Re-init} (CWR) method that skips layers \textit{fc6} and \textit{fc7} of the CNN (for details, see Lomonaco and Maltoni, 2017; page 7), and (iii) a cumulative approach where the learning is carried out by considering the current batch and all the previous ones.

Ours and previously reported experiments show that lifelong learning is a very challenging task and that the overall performance of some approaches can differ significantly according to the specific learning strategy.
Furthermore, a more direct comparison of the model's behavior is hindered by the fact that the other methods do not comprise recurrent neural dynamics that account for learning the temporal structure of the input which, in this case, is a clear advantage (see Table~\ref{Tab:comparison}) since the temporal relations of the input can be exploited for more robust learning and prediction.

The experiments reported for all the 3 incremental learning scenarios were conducted with the test set containing samples from all the seen classes (except for the last evaluation step).
Such an evaluation scheme was selected to keep the test set consistent across all the scenarios~(Lomonaco and Maltoni, 2017). 
However, in a more realistic lifelong learning scenario, the model should be able to deal with unknown classes during sequence retrieval.
In our case, the model will always predict an output label in correspondence to a retrieved sequence.
Instead, it would be convenient to design a novelty detection mechanism for unseen classes so that the system will output a predicted label provided that the input sequence produces a sufficient level of neural activity (Parisi et al., 2015).

\section{Discussion}

\subsection{Summary}

We proposed a growing dual-memory architecture with self-organizing networks for the lifelong learning of spatiotemporal
representations from image sequences. The GDM model
can perform continuous object recognition at an instance
level (episodic experience) and at a category level (semantic
knowledge). We introduced the use of recurrent self-organizing
networks, in particular of extended versions of the GammaGWR, to model the interplay of two complementary learning
systems: an episodic memory, G-EM, with the task of learning
fine-grained spatiotemporal representations from sensory
experience and a semantic memory, G-SM, for learning
more compact representations from episodic experience.
With respect to previously proposed dual-memory learning
systems, our contribution is threefold. First, in contrast
to the predominant approach of processing static images
independently, we implement recurrent self-organizing
memories for learning the spatiotemporal structure of the
input. Second, as a complementary mechanism to unsupervised
growing networks, we use task-relevant signals to regulate
structural plasticity levels in the semantic memory, leading to
the development of more compact representations from episodic
experience. Third, we model memory replay as the periodic
reactivation of neural activity trajectories from temporal synaptic
patterns embedded in an episodic memory. Our experiments
show that the proposed GDM model significantly outperforms
state-of-the-art lifelong learning methods in three different
incremental learning tasks with the CORe50 benchmark
dataset.

\subsection{Growing Recurrent Networks with Memory Replay}

The use of growing networks leads to the dynamic allocation
of additional neurons and connections in response to novel
sensory experience. In particular, the Gamma-GWR (Parisi et al.,
2017) provides the basic mechanism for growing self-organizing
memories with temporal context for learning the spatiotemporal
structure of the input in an unsupervised fashion. Different
models of neural network self-organization have been proposed
that resemble the dynamics of Hebbian learning and plasticity
(Fritzke, 1995; Kohonen, 1995; Marsland et al., 2002), with
neural map organization resulting from unsupervised statistical
learning. For instance, in the traditional self-organizing feature
map and its dynamic variant (e.g., Kohonen, 1995; Rougier
and Boniface, 2011), the number of neurons is pre-defined.

Empirically selecting a convenient number of neurons can be
tedious for networks with recurrent dynamics, especially when
dealing with non-stationary input distributions (Strickert and
Hammer, 2005). To alleviate this issue, growing self-organizing
networks for temporal processing have been proposed, for
instance the Gamma-GNG (Estévez and Vergara, 2012) that
equips neurons with a temporal context. However, the GammaGNG grows at a constant, pre-defined interval and does not
consider whether previously created neurons have been well
trained before creating new ones. This will lead to scalability
issues if the selected interval is too short or, conversely, to
an insufficient number of neurons if the interval is too large.
Therefore, we extended the Gamma-GWR which can quickly
react to changes in the input distribution and can create new
neurons whenever they are required.

From a biological perspective, there has been controversy over
whether in human adults detectable amounts of new neurons
can grow. Recent research has suggested that hippocampal
neurogenesis drops sharply in children (Sorrells et al., 2018)
and becomes undetectable in adulthood, whereas other studies
suggest that hippocampal neurogenesis sustains human-specific cognitive function throughout life (Boldrini et al.,
2018). Neurophysiological studies evidence that, in addition
to neurogenesis, synaptic rewiring by structural plasticity
has a significant contribution on memory formation in
adults (Knoblauch et al., 2014; Knoblauch, 2017), with a major
role of structural plasticity in increasing information storage
efficiency in terms of space and energy demands. While the
mechanisms for creating new neurons and connections in
the Gamma-GWR do not resemble biologically plausible
mechanisms of neurogenesis and synaptogenesis (e.g., Eriksson
et al., 1998; Ming and Song, 2011; Knoblauch, 2017), the GWR
learning algorithm represents an efficient computational model
that incrementally adapts to non-stationary input. Crucially, the
GWR model creates new neurons whenever they are required
and only after the training of existing ones. The neural update
rate decreases as the neurons become more habituated, which
has the effect of preventing that noisy input interferes with
consolidated neural representations. Alternative theories suggest
that an additional function of hippocampal neurogenesis is the
encoding of time for the formation of temporal associations in
memory (Aimone et al., 2006, 2009), e.g., in terms of temporal
clusters of long-term episodic memories. This represents an
interesting research direction for the modeling of temporal
associations in the Gamma-GWR.

For mitigating catastrophic forgetting during incremental
learning tasks, the proposed model generates recurrent
neural activity trajectories (RNATs; Equation 10) after each
learning episode. The set of generated RNATs is periodically
replayed to both networks in correspondence of novel sensory
experience for the consolidation of knowledge over time.
This is consistent with biological evidence suggesting that
the reactivation of hippocampal representations and their
frequent replay to the neocortex are crucial for memory
consolidation and retrieval (see Carr et al., 2011 for a review).
The process of replaying previously seen data without
explicitly storing data samples is referred to as intrinsic
replay (Robins, 1995) and has the advantage of fewer memory
requirements with respect to explicitly storing training samples.
In our approach, the episodic memory G-EM embeds the
temporal structure of the input through the implementation of
temporal synapses that are strengthened between consecutively
activated neurons (Equation 9). Therefore, RNATs comprise
prototype sequence snapshots that can be generated without
the need of explicitly storing the training sequences. Our
reported results show that the use of RNATs yields a
significantly improved overall accuracy during incremental
learning.

In this work, we have focused on regulating the mechanisms
of neurogenesis and neural update, whereas we have not
investigated the removal of old connections and isolated
neurons. At each learning iteration of the Gamma-GWR, old
connections exceeding a given age threshold and neurons
without connections can be deleted. Removing a neuron from
the network means that the knowledge coded by that unit is
permanently forgotten. Therefore, a convenient maximum age
of connections must be set to avoid catastrophic forgetting.
In incremental learning scenarios, it is non-trivial to define
a convenient age threshold for connections to be removed
since data samples become available over time and neurons
coding for consolidated knowledge might not fire for a large
number of iterations. Mechanisms of intrinsic memory replay
as modeled in this paper could be used to prevent the deletion
of consolidated knowledge. For instance, the periodic replay
of episodic representations would prevent the networks from
deleting relevant knowledge also when external sensory input
does not activate those representations for sustained periods of
time.

Conceptual similarities can be found between our model
and the adaptive resonance theory (ART) in which neurons
are iteratively adapted to a non-stationary input distribution
in an unsupervised fashion and new neurons can be added
in correspondence of dissimilar input (see Grossberg, 2012 for
a review). The primary intuition of the ART model is that
learning occurs via the interaction of top-down and bottom-up processes, where top-down expectations act as memory
templates or prototypes which are compared to bottom-up
sensory observations. Similar to the GWR's activation threshold,
the ART model uses a vigilance parameter to produce fine-grained or more general memories. Despite its inherent ability
to mitigate catastrophic forgetting during incremental learning, it
has been noted that the results of some variants of the ART model
depend significantly upon the order in which the training data are
processed. However, an extensive evaluation with recent lifelong
learning benchmarks has not been reported. Therefore, ART-based models represent an additional complementary approach
to growing self-organizing models.

\subsection{Conclusion}

Lifelong learning represents a fundamental but challenging
component of artificial systems and autonomous agents. Despite
significant advances in this direction, current models of lifelong
learning are far from providing the flexibility, robustness,
and scalability exhibited by biological systems. In this paper,
we contribute to extending dual-memory models for the
processing of sequential input which represents more realistic
experimental settings compared to learning from static image
datasets. In the future, it would be interesting to extend
this model to the multisensory domain, e.g., where neural
representations can be continually learned from audio-visual
streams (Parisi et al., 2016, 2018b). The proposed architecture
can be considered a further step toward more flexible lifelong
learning methods that can be deployed in embodied agents
for incrementally acquiring and refining knowledge over
sustained periods through the active interaction with the
environment.

\section*{Conflict of Interest Statement}

The authors declare that the research was conducted in the absence of any commercial or financial relationships that could be construed as a potential conflict of interest.

\section*{Acknowledgments}
This research was partially supported by the German Research Foundation (DFG) under project Transregio Crossmodal Learning (TRR 169).
The authors would like to thank Pablo Barros and Vadym Gryshchuk for technical support.



\bibliographystyle{frontiersinSCNS_ENG_HUMS} 

\end{document}